\documentclass[conference, final]{IEEEtran}
\IEEEoverridecommandlockouts
\usepackage{cite}
\usepackage{amsmath,amssymb,amsfonts}
\usepackage{breqn}
\usepackage{algorithmic}
\usepackage{graphicx}
\usepackage{textcomp}
\usepackage{xcolor}
\usepackage{tikz}
\usepackage{multirow}
\def\BibTeX{{\rm B\kern-.05em{\sc i\kern-.025em b}\kern-.08em
    T\kern-.1667em\lower.7ex\hbox{E}\kern-.125emX}}

\newcommand{\vx}{\mathbf{x}}
\newcommand{\vb}{\mathbf{b}}
\newcommand{\va}{\mathbf{a}}
\newcommand{\vz}{\mathbf{z}}
\newcommand{\vy}{\mathbf{y}}

\IEEEoverridecommandlockouts

\usepackage{tikz}
\usepackage{textcomp}
\usepackage{hyperref}
\usepackage{lipsum}

\newcommand\copyrighttext{%
  \footnotesize \textcopyright 2021 IEEE. Personal use of this material is permitted.
  Permission from IEEE must be obtained for all other uses, in any current or future
  media, including reprinting/republishing this material for advertising or promotional
  purposes, creating new collective works, for resale or redistribution to servers or
  lists, or reuse of any copyrighted component of this work in other works.}
\newcommand\copyrightnotice{%
\begin{tikzpicture}[remember picture,overlay]
\node[anchor=south,yshift=10pt] at (current page.south) {\fbox{\parbox{\dimexpr\textwidth-\fboxsep-\fboxrule\relax}{\copyrighttext}}};
\end{tikzpicture}%
}

\begin{document}

\title{Volume-preserving Neural Networks\\
\thanks{The first author acknowledges the support of NSERC Canada. The third and fourth author acknowledge the support of NSERC Canada through the NSERC USRA program.}
}

\author{\IEEEauthorblockN{Gordon MacDonald\IEEEauthorrefmark{1}, Andrew Godbout\IEEEauthorrefmark{1}, Bryn Gillcash\IEEEauthorrefmark{1} and Stephanie Cairns\IEEEauthorrefmark{4}}
\IEEEauthorblockA{\IEEEauthorrefmark{1}\textit{School of Mathematical and Computational Sciences,}
\textit{University of Prince Edward Island,}
Charlottetown, Canada \\
\{gmacdonald@upei.ca, agodbout@upei.ca, bwgillcash@upei.ca\}}
\IEEEauthorblockA{\IEEEauthorrefmark{4}\textit{Department of  Mathematics and Statistics,}
\textit{McGill University,}
Montreal, Canada \\
stephanie.cairns@mail.mcgill.ca}
}

\maketitle
\copyrightnotice
\begin{abstract}
We propose a novel approach to addressing the vanishing (or exploding) gradient problem in deep neural networks. We construct a new architecture for deep neural networks where all layers (except the output layer) of the network are a combination of rotation, permutation, diagonal, and activation sublayers which are all volume preserving. Our approach replaces the standard weight matrix of a neural network with a combination of diagonal, rotational and permutation matrices, all of which are volume-preserving. We introduce a coupled activation function allowing us to preserve volume even in the activation function portion of a neural network layer. This control on the volume forces the gradient (on average) to maintain equilibrium and not explode or vanish. To demonstrate our architecture we apply our volume-preserving neural network model to two standard datasets.
\end{abstract}


\section{Introduction}

Deep neural networks are characterized by the composition of a large number of functions (aka layers), each typically consisting of an affine transformation followed by a non-affine ``activation function''. Each layer is determined by a  number of parameters which are trained on data to approximate some function. The deepness refers to the number of such functions composed (or the number of layers). The number of layers required to be deep is not well-defined, but an overview of deep learning \cite{S2} states that any network with more than three layers is deep, and any network with more than ten layers is very deep.

Deep neural networks have been successfully applied to a number of difficult machine learning problems, such as image recognition \cite{KSH}, speech  recognition \cite{Hetal}, and natural language processing \cite{Cetal}. In deep neural networks trained via gradient descent methods with backpropagation, the problem of vanishing gradients \cite{H}, \cite{BSF}, \cite{glorot2010understanding}  makes it difficult to train the parameters of the network. The backpropagation equations, via the chain rule, multiply a large number of derivatives in deep networks. If too many of these derivatives are small, the gradients vanish, and little learning happens in early layers of the network. In the standard neural network model, there are two main contributors to small derivatives: activation functions which often squash vectors and as such have small derivatives on a large portion of their domain; and weight matrices which act compressively on large parts of their domain.

There have been a number of approaches to address the vanishing (or exploding gradient) problem. These techniques include using   alternative activation functions (such as ReLU) \cite{nair2010rectified}, clipping gradients \cite{pascanu2013difficulty}, long short-term memory (LSTM) units \cite{HS},  gated recurrent units (GRU)  \cite{Cetal}, multi-level hierarchies \cite{S1} and orthogonal constraints on parameter initializations \cite{saxe2013exact}.

There are a number of approaches that focus on adjustments to the standard weight matrices of the network such as the use of orthogonal or unitary constraints \cite{bansal2018can,henaff2016recurrent,arjovsky2016unitary,wisdom2016full}. The approaches in \cite{arjovsky2016unitary} and \cite{wisdom2016full} enforce the weight matrices within a recurrent neural network to be unitary and achieve this by parameterizing the weight matrix as a product of unitary building blocks (such as a combination of diagonal, permutation and reflection matrices). Although these approaches are applied to recurrent neural networks they share similarities with our approach and especially in their parameterization of the weight matrix to ensure it is unitary.




Our approach proposes adjustments to the the activation functions and the weight matrices, by replacing each of them with mathematical variants which are volume preserving.  Enforcing  volume preservation ensures that gradients cannot universally vanish or explode. We replace the standard weight matrix with a product of rotation, permutation, and diagonal matrices, all of which are volume preserving.  We replace the usual entrywise-acting activation functions by coupled activation functions which act pairwise on entries of an input vector (rather than entrywise) and allows us to use a wider selection of activation functions, ones that can ``squash'' while still being volume preserving.




The width of a network is the maximum number of inputs accepted to a layer of the network. If the width $w$ is roughly constant across a network $d$ layers deep, the number of parameters required in  a standard fully-connected network is of order $w^2  d$.  In our VPNN, the number of parameters required is  of order $w \log_2( w) \, d $, yet, despite this significant reduction in the number of parameters as compared to a fully-connected network our results are similar in our testing.


Since being \emph{volume preserving} is at the core of the model, we begin by reminding the reader of the definition. A function $f:\mathbb{R}^n \rightarrow \mathbb{R}^n$ is volume-preserving if
$\hbox{vol}( f^{-1}(S) ) = \hbox{vol}(S)$  for all  measurable sets $S \subset \mathbb{R}^n$
(where $\hbox{vol}(\cdot)$ is the usual (Lebesgue) volume of a set).

\section{The Basic VPNN Model}\label{model}
The basic $L$ layer VPNN will take $n_{in}$ inputs, process them through  $L-1$  volume-preserving layers (the input layer and the hidden layers) and an output layer to produce $n_{out}$ outputs. Each volume-preserving layer (for $l= 1,2,\ldots, L-1$ is of the form:
\begin{equation}\mathbf{x}\rightarrow A(V^{(l)} \mathbf{x} + \mathbf{b}^{(l)})\end{equation}
 where $V^{(l)}$ is a volume-preserving linear transformation,  $\vb^{(l)}$ is a bias vector, and $A$ is a volume-preserving coupled activation function.

Being volume preserving necessarily implies being dimension preserving, so in $L-1$ volume-preserving layers $V^{(l)}$ is an $n_{in} \times n_{in}$ matrix, $\vb^{(l)}$ is a vector in $\mathbb{R}^{n_{in}}$, and $A$ is a function from $\mathbb{R}^{n_{in}}$ to itself.

The $L$-th layer (the output layer) is necessarily not volume preserving as it must downsize to the size of the classifier space. In the basic VPNN we implement this by a fixed $n_{out}\times n_{in} $ matrix $Z$ so the output layer is just:
\begin{equation}\vx\rightarrow Z \vx. \end{equation}

\subsection{Building the Volume-Preserving Linear Transformations of a VPNN}

We build $V$, a volume-preserving linear transformation, as a product of rotation, permutation, and diagonal matrices. We first describe in detail these matrices and then describe how we fit them together in the VPNN architecture.

Let:
\begin{equation}
R_\theta = \begin{bmatrix} \cos \theta & -\sin \theta \\ \sin \theta &\cos \theta \end{bmatrix}
\end{equation}
(the matrix that rotates a vector in $\mathbb{R}^2$ by $\theta$ in the counterclockwise direction). Then a rotation matrix $R$ in a VPNN corresponds to a direct sum of such matrices:
\begin{equation} R = \bigoplus_{i=1}^{n_{in}/2} R_{\theta_i} =  \begin{bmatrix}   R_{\theta_1 } &  0   & 0 & \cdots& & 0 \\
                                                                                            0  &  R_{\theta_2 }   &0 & \cdots&& 0 \\
                                                                                            0 & 0 & R_{\theta_3 } &     & &  \\
                                                                                            \vdots& \vdots &&\ddots&&\\
                                                                                            0&  0  & & & & R_{\theta_{n_{in}/2} } \end{bmatrix}.
                                                                                            \end{equation}
There are  $n_{in}/2$ trainable parameters in a rotation matrix, each parameter is involved in four neuron connections and each input neuron connects to two output neurons.

A permutation matrix $Q$  in a VPNN corresponds to  a permutation $q$ of $\{1,2,3, \ldots,n_{in}\}$ (a bijection from  $\{1,2,3, \ldots,n_{in}\}$ to itself) which is chosen randomly before training begins. So the permutation matrix $Q$ has  $(q(i),i)$ entries (for $i =1,2,\ldots n_{in}$) equal to one and  all other entries are zero. There are no trainable parameters in a permutation matrix, and each input neuron connects to one output neuron.

A diagonal matrix $D$  in a VPNN has  diagonal entries which are positive and have product one.
To stay away from possible ``division by zero'' problems,  we  implement this as
\begin{equation}
D= \begin{bmatrix}  \frac{f(t_1)}{f(t_{n_{in}})} &  & & &   \\
                                            &  \frac{f(t_2)}{f(t_1)} &&&  \\
                                          &                               & \ddots && \\
                                           &&&                                             \frac{ f(t_{n_{in}-1}) } {f(t_{n_{in}-2})} & \\
                                                                           &&&& \frac{f(t_{n_{in}})}{ f(t_{n_{in}-1}) }
                                                                           \end{bmatrix}
                                                                           \end{equation}
where $f$ is a function from $\mathbb{R}$ to $\mathbb{R}^+$ whose range lies in some compact interval (and all off-diagonal entries are zero). In our implementation we choose  $f(x) = \exp (\sin x) $.

There are $n_{in}$ trainable parameters in each diagonal matrix, each parameter is involved in two neuron connections and each input neuron connects to one output neuron.


Then the volume-preserving linear transformation $V$ is implemented as
\begin{equation}
 V= \left( \prod_{j=1}^{k/2} R_j Q_j \right) D \left( \prod_{j=k/2 +1}^{k} R_j Q_j \right)
 \end{equation}
with each $R_jQ_j$ connecting two input neurons to two ``random'' output neurons, using $\left\lceil \log_2 (n_{in}) \right\rceil$ such $R_jQ_j$ along with a diagonal matrix should achieve almost total neuronal interaction in each volume-preserving affine layer.  However, testing showed  there is a slight improvement in accuracy when we add additional $R_jQ_j$ to gain some redundant neural connections. So, in the basic VPNN model we set $k$ (the number of rotations or permutations used in any layer) to be:
\begin{equation}
k=  2  \left\lceil \log_2 (n_{in}) \right\rceil.
 \end{equation}
 This also ensures $k$ is even and so allows us to have the same number of rotations and permutations on each side of the diagonal (this is not strictly necessary). Not surprisingly, the more layers in the VPNN, the less pronounced is the effect of adding redundant rotations/permutations in any layer. In very deep networks,  taking $k$ closer to $\left\lceil \log_2 (n_{in}) \right\rceil$ is probably optimal.

\subsection{Building the Coupled Activation Functions of a VPNN}\label{couple}

A coupled activation function corresponds to a non-affine function $C$  from $\mathbb{R}^2$ to $ \mathbb{R}^2$ which is area preserving. Instead of the usual activation functions, which act  entrywise on the entries of a vector, a coupled activation function $A$ acts on a vector $\vx$ (with an even number of entries) by grouping them in pairs and applying $C$ to them pairwise. So a coupled activation sublayer performs:
\begin{equation}
\vx = \begin{bmatrix} x_1 \\ x_2 \\ \vdots \\ \vdots \\ x_{n-1} \\ x_n\end{bmatrix} \xrightarrow[]{A} \begin{bmatrix} C \left( \begin{bmatrix} x_1 \\ x_2 \end{bmatrix} \right) \\ \vdots \\ \vdots \\ C \left( \begin{bmatrix} x_{n-1} \\ x_n \end{bmatrix} \right) \end{bmatrix}.
\end{equation}
Such functions can be created in many ways, but for our basic VPNN model we use what we refer to as  \emph{coupled Chebyshev functions}.

These functions are most easily described in polar coordinates. Given a point $(x,y)$ in the plane, if $r$ is the distance from that point to $(0,0)$ and  $-\pi < \theta \leq \pi$ is the angle the ray from $(0,0)$ to $(x,y)$ makes with the positive $x$ axis, then  $r=\sqrt{x^2+y^2}$ and $\theta = \hbox{sgn}(y) \cos^{-1} \left( \frac{x}{\sqrt{x^2+y^2}} \right)$ are the polar coordinates of $(x,y)$. We introduce a contractive factor $M$ and define:
\begin{equation}
C_M(r, \theta) = \left( \frac{r}{\sqrt{M} }, M\theta \right)
\end{equation}
so the radius $r$ is contracted by $\sqrt{M} $ and the angle $\theta$ is increased by a factor of $M$. The area unit for polar coordinates is $dA= r \, dr \, d\theta $ so:
\begin{align}
d(C_M(A)) & = \frac{r}{\sqrt{M}} \frac{\partial C_M}{\partial r} \frac{\partial C_M}{\partial \theta} dr \, d\theta \\
&= \frac{r}{\sqrt{M} }\frac{1}{\sqrt{M}} M\, dr \, d\theta = r \, dr \, d\theta.\end{align}
Converting to Cartesian coordinates,
\begin{align}
C_M \left( \begin{bmatrix} x \\ y\end{bmatrix} \right)= \begin{bmatrix} \frac{\sqrt{x^2+y^2}}{\sqrt{M}} \cos\left( M \cos^{-1} \left( \frac{x}{\sqrt{x^2+y^2}} \right)\right) \\
  \frac{\sqrt{x^2+y^2}}{\sqrt{M}} \hbox{sgn}(y) \sin\left( M \cos^{-1} \left( \frac{x}{\sqrt{x^2+y^2}} \right) \right) \end{bmatrix}.
  \end{align}
This is the formula we will be using in our coupled activation function, typically with a value of $M$ in the range $(1,2]$. Just for interest we mention that, in the case where $M$ is an even integer, these are related to the famous Chebyshev polynomials:
\begin{equation}C_M \left( \begin{bmatrix} x \\ y\end{bmatrix} \right)= \begin{bmatrix} \frac{\sqrt{x^2+y^2}}{\sqrt{M}}  T_M \left(  \frac{x}{\sqrt{x^2+y^2}} \right) \\
    \frac{ |y|}{\sqrt{M}} U_{M-1}\left(  \frac{x}{\sqrt{x^2+y^2}} \right) \end{bmatrix}
\end{equation}
where $T_n$ is Chebyshev polynomial of the first kind ($T_n(x) = \cos(n\cos^{-1}(x))$), and $U_{n}$ is Chebyshev polynomial of the second kind ($U_n(x) = \frac{\cos(n\sin^{-1}(x))}{\sin(\cos^{-1} (x) )}$).
In the case $M=2$ these have particularly nice form:
\begin{equation}
C_2 \left( \begin{bmatrix} x \\ y\end{bmatrix} \right) = \begin{bmatrix} \frac{x^2-y^2}{\sqrt{2}\sqrt{x^2+y^2} } ,    \frac{\sqrt{2}x |y|}{\sqrt{x^2+y^2}} \end{bmatrix}.
\end{equation}

\subsection{Building the Output Layer of a VPNN}

Since volume-preserving layers cannot downsize (reduce dimension) we need some method to map down to the dimension of our classification space. We could use a fully-connected layer, but in the testing that follows we want to demonstrate that the learning is happening in the volume-preserving layers, so our output layer will have no trainable parameters.

We implement this as simply as possible. We use no bias on this layer and fix a  ``random'' matrix $Z$ of size $n_{out} \times n_{in}$  with $ZZ^T=1$ and  with most entries non-zero and of roughly equal magnitude.
 (This is chosen to preserve length and connect every output neuron in this layer to every input neuron with roughly the same weight).  Then the output layer performs:
\begin{equation}
\vx \rightarrow Z\vx.\end{equation}

So, roughly, we are just choosing a random initialization of a weight matrix $Z$, but not allowing the weights to train in this final layer.

We generate the downsizer matrix $Z$ by randomly choosing entries of an $n_{out} \times n_{in}$ matrix $A$ from the interval $[-1,1]$, then applying the reduced Singular Value Decomposition to $A$, we obtain so $A = U \Sigma V^T$  where  $Z=U$ has the desired properties.


\section{Discussion of the VPNN Model}\label{feat}

The key feature of our neural network is that it is volume preserving in all layers except the output layer. Rotations, permutations,  and translations are rigid maps on $\mathbb{R}^n$ and so leave volume unchanged. The determinant one condition ensures the diagonal layer is also volume preserving, and the coupled activation maps are also volume preserving.
Because of the volume-preserving property, if vectors on the unit ball are sent through a  layer, some will be shortened and some lengthened. When composing through multiple hidden layers, we would expect ``on average'' that a vector will be shortened at some layers and lengthened at others and generally not have its length vanish or explode, thus giving some management of the gradient.

Once being volume preserving was identified as a control mechanism for the gradient, we needed volume-preserving activation functions. Since activation functions are necessarily non-affine, they cannot be constructed as functions of one input variable only. So we had to allow coupled activation functions which take two (or more) inputs.

Next we needed finer control on the weights layer. Our construction is motivated by the Singular Value Decomposition, which states that any square matrix can be written as $UDV$ where $U$ and $V$ are orthonormal (i.e. real unitary) and $D$ is diagonal with non-negative diagonal entries. Any real unitary matrix (of determinant 1) can be written as a product of Givens rotations. Every Givens rotation is of the form $QRQ^{-1}$ for some permutation matrix and some choice of parameters $\theta_i$ (all but one chosen to be zero). Thus it is reasonable that we should be able to replace a general weight matrix  $W$ by a volume-preserving matrix $V$ of the above form with  little impact on ability to approximate.



The number of trainable parameters in each of the first $L-1$ layers of a basic VPNN is $n_{in} (\lceil \log_2 n_{in} \rceil + 2)$ where $n_{in}$ is the number of entries in the input vector to the neural network ( $n_{in} \lceil \log_2 n_{in} \rceil$ from rotations, $n_{in}$ from diagonals, and $n_{in}$ from biases).  Contrast this to $n_{in}^2 +n_{in}$ in a standard neural network (or even greater if there was upsizing).


\begin{figure}
\centering
\resizebox{\columnwidth}{!}{%
\begin{tikzpicture}[shorten >=1pt,->,draw=black!50, node distance=2cm]
    \tikzstyle{every pin edge}=[<-,shorten <=1pt]
    \tikzstyle{neuron}=[circle,fill=black!25,minimum size=15pt,inner sep=0pt]
    \tikzstyle{layer0 neuron}=[neuron, fill=green!50];
    \tikzstyle{layer1 neuron}=[neuron, fill=red!50];
    \tikzstyle{layer2 neuron}=[neuron, fill=blue!50];
     \tikzstyle{layer3 neuron}=[neuron, fill=purple!50];
      \tikzstyle{layer4 neuron}=[neuron, fill=yellow!50];
    \tikzstyle{annot} = [text width=4em, text centered]

    \foreach \name / \y in {1,...,4}
        \node[layer0 neuron] (L0-\name) at (0,-\y) {};

    \foreach \name / \y in {1,...,4}
        \path
            node[layer1 neuron] (L1-\name) at (2.5cm,-\y cm) {};

    \foreach \name / \y in {1,...,4}
        \path
            node[layer2 neuron] (L2-\name) at (5cm,-\y cm) {};

    \foreach \name / \y in {1,...,4}
        \path
            node[layer3 neuron] (L3-\name) at (7.5cm,-\y cm) {};

    \foreach \name / \y in {1,...,4}
        \path
            node[layer4 neuron] (L4-\name) at (10cm,-\y cm) {};


    \foreach \source in {1,...,4}
            \path (L0-\source) edge (L1-\source);


            \path (L1-1) edge (L2-3);
            \path (L1-2) edge (L2-1);
            \path (L1-3) edge (L2-2);
            \path (L1-4) edge (L2-4);

    \foreach \source in {1,...,2}
        \foreach \dest in {1,...,2}
            \path (L2-\source) edge (L3-\dest);

        \foreach \source in {3,...,4}
        \foreach \dest in {3,...,4}
            \path (L2-\source) edge (L3-\dest);

    \foreach \source in {1,...,2}
        \foreach \dest in {1,...,2}
            \path (L3-\source) edge (L4-\dest);

        \foreach \source in {3,...,4}
        \foreach \dest in {3,...,4}
            \path (L3-\source) edge (L4-\dest);

    \node[annot,below of=L1-3] {{\hskip -1.5cm} Diagonal};
     \node[annot,below of=L2-3] {{\hskip -1.5cm} Permutation};
      \node[annot,below of=L3-3] {{\hskip -1.5cm} Rotation};
       \node[annot,below of=L4-3] {{\hskip -1.5cm} Activation};
\end{tikzpicture}
}
  \caption{A simplified VPNN layer}\label{v}
\end{figure}
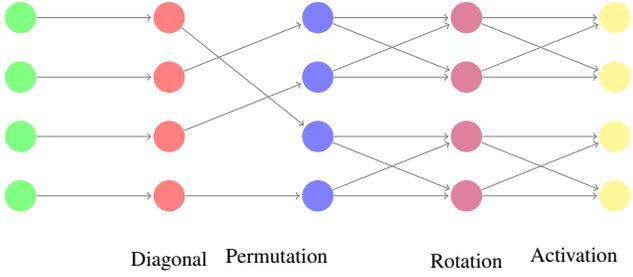

In \cite{GBB},  they illustrate the superior performance of  ReLU compared to other activation functions in deep neural networks. One possible explanation for this superior performance, as mentioned in the paper, is the fact that ReLU introduces \emph{sparsity}. Certain neuronal connections are pruned by virtue of having negative inputs into ReLU. In a VPNN, this sparsity is incorporated by a different mechanic, not by pruning but by building fewer neuronal connections as part of the architecture.

We scale our input vectors so that their length is within the same order of magnitude of the output vectors (which should have length 1, if the network is learning correctly). In practice we preprocess our inputs by scaling entries so that they lie in some interval (say $[0,1]$) and then divide each entry by $\sqrt{n_{in}}$ where $n_{in}$ is the number of entries, so that the length of an input vector is reasonably close to 1.  This is often done in neural network models but is particularly important for VPNNs since any stretching or compressing in the basic VPNN must be done in diagonal and activation layers, and we do not want to impose extra work on these layers to scale vectors up or down  beyond what is needed for approximation.

The VPNN must also  act on vectors with an even number of entries, as the rotational layers and coupled activation layers require an even number of inputs. If we had an odd number of inputs the simplest solution would be to add one new input which was always zero.

\section{Backpropagation Equations}\label{back}


Let $\theta_{p,i}^{(l)}$ denote the $i^{\text{th}}$ rotational parameter ($i=1,2,\ldots n_{in}/2$) in the $p^{\text{th}}$ rotation matrix ($p=1,2,\ldots ,k$) in the $l^{\text{th}}$ layer ($l=1,2,\ldots , L-1$) and let $t_j^{(l)}$ denote the $j^{\text{th}}$ diagonal parameter in the diagonal matrix $D^{(l)}$ in the $l^{\text{th}}$ layer ($l=1,2,\ldots,  L-1$), and let $b_j^{(l)} $ denote the $j^{\text{th}}$ bias parameter in the bias vector $\vb^{(l)}$ in the $l^{\text{th}}$ layer ($l=1,2,\ldots ,L-1$)

For a given error function (or cost function) $E$, we need to compute:
for all bias sublayers: $\frac{\partial E}{\partial b_j^{(l)} }$ for $l=1,2,\cdots , L-1,$ for all rotational sublayers: $\frac{\partial E}{\partial \theta_{p,i}^{(l)} }$  for $l=1,2,\cdots , L-1$, and for all diagonal sublayers: $\frac{\partial E}{\partial t_j^{(l)} }$ for $l=1,2,\cdots , L-1.$\\

For a single $\vx_{in}= \va^{(0)}$ sent through the network generating output $\vy_{out} = \va^{(L)}$, we use the following terminology for partially forward-computed terms:
\begin{align}
&V_{\text{left}}^{(l)}  = \prod_{j=1}^{k/2} R_{j}^{(l)} Q_j^{(l)}, \\
&V_{\text{right}}^{(l)} = \prod_{j=k/2 +1}^{k }R_{j}^{(l)} Q_j^{(l)},\\
&V^{(l)}  = V_{\text{left}}^{(l)} D^{(l)}  V_{\text{right}}^{(l)}  \\
&\vz^{(l)} = V^{(l)} \va^{(l-1)} + \vb^{(l)}   \quad \hbox {for } l=1,2,\ldots , L-1  \\
&\va^{(l)} = A( \vz^{(l)} )   \hskip .5 true in \hbox{for } l=1,2,\ldots , L-1 \\ &\va^{(L)} = Z \va^{(L-1)}.  \\
 \end{align}

 Define (for $l=1,2,\ldots, L-1$)
\begin{equation}
 \delta^{(l)} = \frac{\partial E }{\partial \vz^{(l)} } =\begin{bmatrix}  \frac{\partial E }{\partial z_1^{(l)} } \\  \frac{\partial E }{\partial z_2^{(l)} } \\ \vdots \\\frac{\partial E }{\partial z_{n_{in}}^{(l)} } \end{bmatrix}.
 \end{equation}

 Then we have the following backpropagation equations to backpropagate completely through a layer. For any coupled activation function as described above:

\noindent For $l=1,2,3,\ldots,L-1$:

\noindent If $j$ is odd:
\begin{align}
 \delta_j^{(l)}  = &
 \left( {V^{(l)} }^T \delta^{(l+1)} \right)_j \left( \frac{\partial C_1}{\partial x} \Bigg\vert_{
 \tiny{ \begin{matrix} x= z_j^{(l)} \\  y = z_{j+1}^{(l)} \end{matrix}  }  }  \right)
 + \nonumber\\
 & \left( {V^{(l)} }^T \delta^{(l+1)} \right)_{j+1} \left( \frac{\partial C_2}{\partial x} \Bigg\vert_{\tiny{ \begin{matrix} x= z_j^{(l)} \nonumber\\  y = z_{j+1}^{(l)} \end{matrix}  } } \right)  \\
\end{align}

\noindent If j is even:\\

\begin{align}
\delta_j^{(l)}  = & \left( {V^{(l)} }^T \delta^{(l+1)} \right)_{j-1} \left( \frac{\partial C_1}{\partial y} \Bigg\vert_{
\tiny{ \begin{matrix} x= z_{j-1}^{(l)} \\  y = z_{j}^{(l)} \end{matrix}  }  } \right)
+ \nonumber\\
& \left(  {V^{(l)} }^T \delta^{(l+1)} \right)_{j} \left( \frac{\partial C_2}{\partial y} \Bigg\vert_{\tiny{ \begin{matrix} x= z_{j-1}^{(l)}  \\  y = z_{j}^{(l)} \end{matrix}  } } \right)
\end{align}
(where $C_1(x,y)$ is the first component of the coupled activation function and $C_2(x,y)$ is the second component).

  In the case of the coupled Chebyshev activation function, these partials simplify quite nicely in terms of previously computed quantities.
 \begin{align*}
 \frac{\partial C_1}{\partial x} &= \frac{1}{x^2+y^2} \begin{bmatrix} x &  My \end{bmatrix}   C\left( \begin{bmatrix} x \\ y \end{bmatrix} \right)  \\
  \frac{\partial C_2}{\partial x} &=\frac{1}{x^2+y^2} \begin{bmatrix} -My &  x \end{bmatrix}  C\left( \begin{bmatrix} x \\ y \end{bmatrix} \right)    \\
  \frac{\partial C_1}{\partial y} &= \frac{1}{x^2+y^2} \begin{bmatrix} y &  -Mx \end{bmatrix}  C\left( \begin{bmatrix} x \\ y \end{bmatrix} \right)   \\
  \frac{\partial C_2}{\partial y} &= \frac{1}{x^2+y^2} \begin{bmatrix} M x &  y \end{bmatrix}  C\left( \begin{bmatrix} x \\ y \end{bmatrix} \right) .
  \end{align*}

 The following equations allow us to backpropagate through sublayers of a layer.
In all (non-output) layers $l=1,2,\ldots , L-1$ the bias parameters have partials:
\begin{equation}\frac{\partial E}{\partial b_{j}^{(l)} } = \delta_j^{(l)}. \end{equation} In the (non-output) layers  $l=1,2,\ldots , L-1$ the diagonal parameters have partials:
 \begin{equation}
  \frac{\partial E}{\partial t_{j}^{(l)}  }= {\delta^{(l)} }^T \left(V_{\text{left}}^{(l)} F_j^{(l)}  V_{\text{right}}^{(l)}  \right) a^{(l-1)},
\end{equation}
 where
 $F_j^{(l)}$ is a diagonal matrix (of same size as $D^{(l)}$) whose $j$-th diagonal entry is $f^\prime ( t_{j}^{(l)} )$ and whose $j+1$-th diagonal entry (modulo $n_{in}$) is $-\frac{f(t_{j+1}^{(l)}) }{ f(t_{j}^{(l)} )^2 } f^\prime ( t_{j}^{(l)} )$ and all other diagonal entries are zero.
In the layers $l=1,2,\ldots , L-1$, the rotational parameters have partials, for $p=1,2, \ldots k/2$ then:
 \begin{align}
&\frac{\partial E}{\partial \theta_{p,i}^{(l)} } =\nonumber\\
&{\delta^{(l)}}^T  \left(\prod_{j=1}^{p-1} R_{j}^{(l)} Q_j^{(l)} \right) Z_{i} \left(\prod_{j=p}^{k/2} R_{j}^{(l)} Q_j^{(l)}\right)D^{(l)}  V_{\text{right}}^{(l)}  a^{(l-1)}.
\end{align}
For $p= k/2 +1, \ldots k$ then:
\begin{align}
&\frac{\partial E}{\partial \theta_p^{(l)} } =\nonumber\\
&{\delta^{(l)}}^T V_{\text{left}}^{(l)}  D^{(l)}  \left( \prod_{j=k/2 +1}^{p-1 }R_{j}^{(l)} Q_j^{(l)}\right) Z_{i} \left( \prod_{j=p}^{k}R_{j}^{(l)} Q_j^{(l)} \right)a^{(l-1)}
\end{align}
where $Z_i$ is the matrix with a $1$ in the $(2i-1, 2i)$ entry, a $-1$ in the $ (2i,2i-1) $ entry and all other entries are zero. (Note this is equivalent to inserting into the formula for $V^{(l)}$, before the location of rotation $p$, a new matrix which has a $ 2 \times 2$ rotation matrix $R_{\pi/2}$ in the block corresponding to parameter $\theta_{p,i}$ and zeroes elsewhere.)

 \subsection{Variant: Trainable parameters in Coupled Activation Sublayers}\label{Ct}

 It is not much more costly to allow the parameters $M$ in  Coupled Activation Sublayers using  coupled Chebyshev functions to be trainable. In fact we could implement the Coupled Activation Layer as:
\begin{equation}
\vx = \begin{bmatrix} x_1 \\ x_2 \\ \vdots \\ \vdots \\ x_{n-1} \\ x_n\end{bmatrix} \xrightarrow[]{A} \begin{bmatrix} C_{M_1} \left( \begin{bmatrix} x_1 \\ x_2 \end{bmatrix} \right) \\ \vdots \\ \vdots \\ C_{M_{n/2}} \left( \begin{bmatrix} x_{n-1} \\ x_n \end{bmatrix} \right) \end{bmatrix}
\end{equation}
where $M_1, M_2, \ldots , M_{n/2}$ are $n/2$ trainable parameters.

The derivatives with respect to $M_i$ to be used in the modified backpropagation equations are:
\begin{equation}
\frac{d}{dM_i} C_{M_i} \left(\begin{bmatrix} x \\ y\end{bmatrix} \right) = \begin{bmatrix}  - \frac{1}{2M_i} &  - \theta_{(x,y)}   \\  \theta_{(x,y)}  & - \frac{1}{2M_i}  \end{bmatrix} C_{M_i} \left(\begin{bmatrix} x \\ y\end{bmatrix} \right)
\end{equation}
where:
\begin{equation}\theta_{(x,y)} = \hbox{sgn}(y)\cos^{-1}\left( \frac{x}{\sqrt{x^2 + y^2}} \right). \end{equation}

\section{Testing} \label{test}
\begin{figure}
   \centering
   \begin{minipage}{0.47\textwidth}
       \centering
       \includegraphics[width=1.1\textwidth]{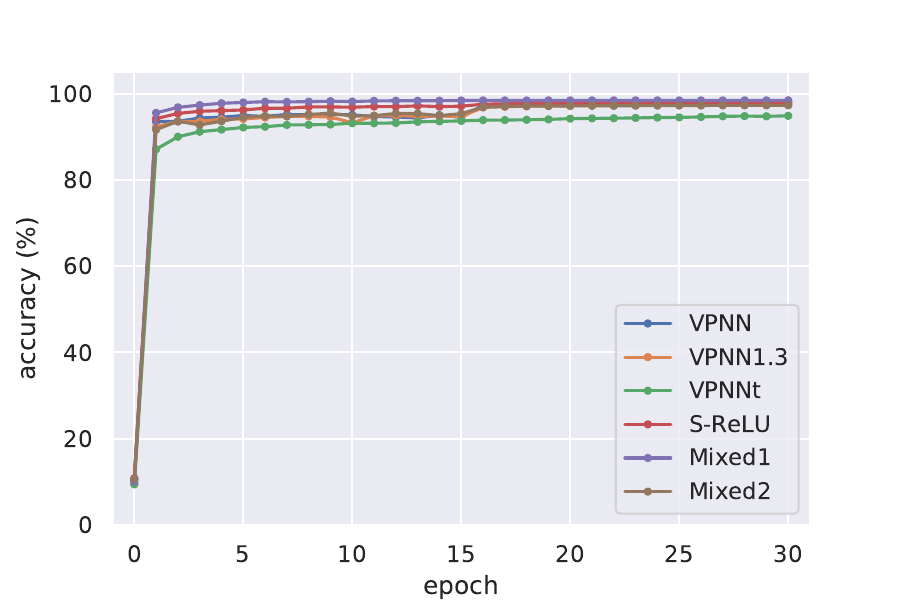} 
       \caption{Accuracy:MNIST}\label{mnist-a}
   \end{minipage}\hskip .03\textwidth
   \begin{minipage}{0.47\textwidth}
       \centering
       \includegraphics[width=1.1\textwidth]{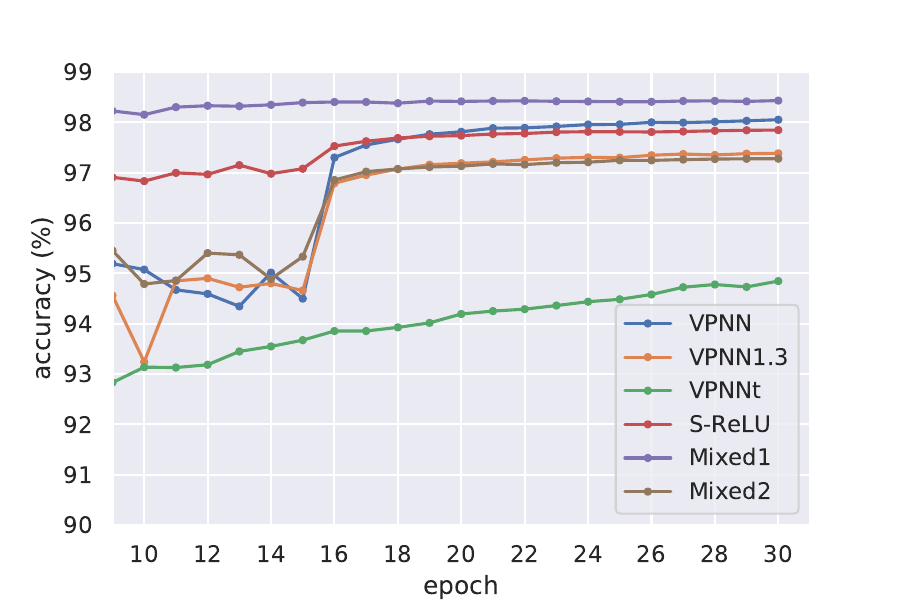} 
       \caption{Zoomed Accuracy:MNIST}\label{mnist-a-z}
   \end{minipage}
\end{figure}

\begin{figure}
   \centering
   \begin{minipage}{0.47\textwidth}
       \centering
       \includegraphics[width=1.1\textwidth]{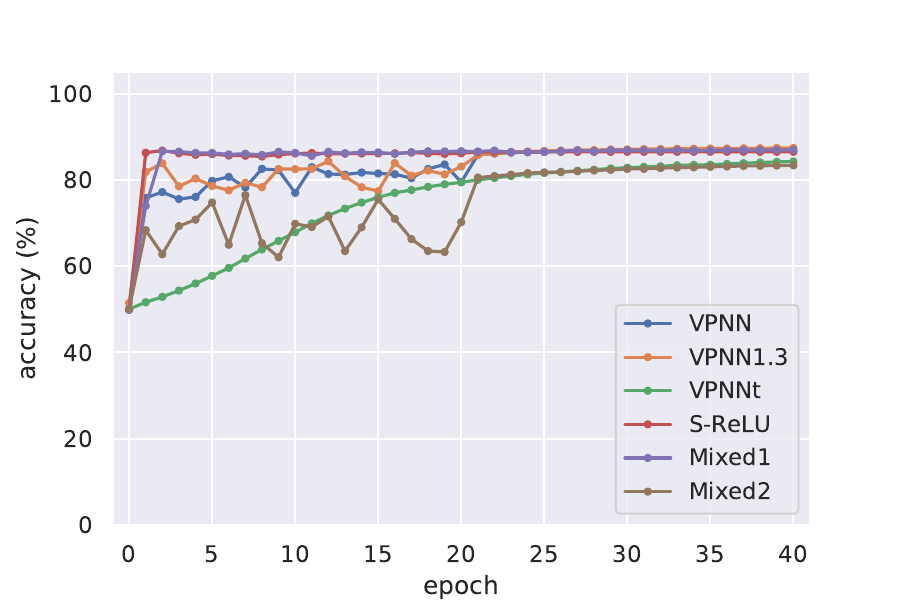} 
       \caption{Accuracy:IMDB}\label{imdb-a}
   \end{minipage}\hskip .03\textwidth
   \begin{minipage}{0.47\textwidth}
       \centering
       \includegraphics[width=1.1\textwidth]{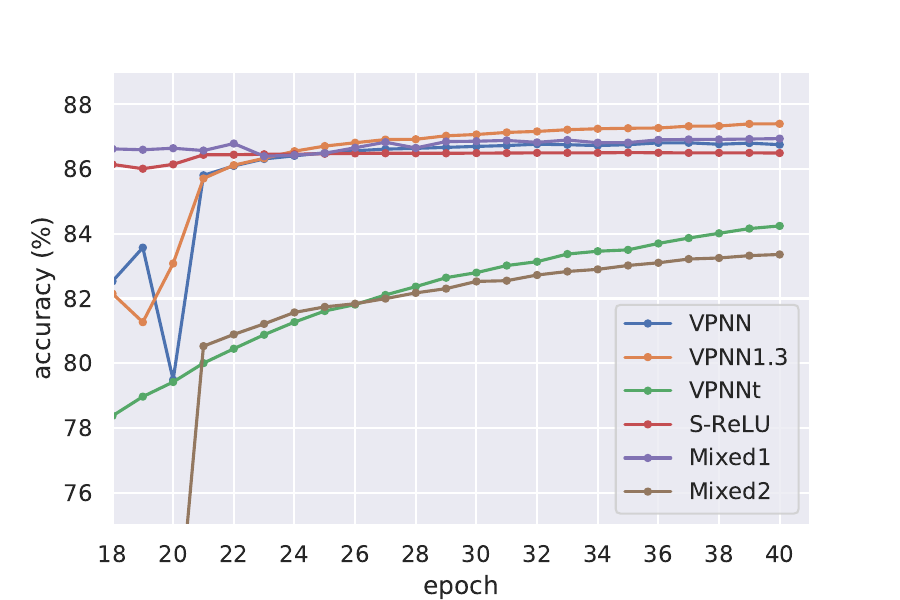} 
       \caption{Zoomed Accuracy:IMDB}\label{imdb-a-z}
   \end{minipage}
\end{figure}

To demonstrate the utility of VPNN, we compare its performance in terms of accuracy, training time, and size of gradients throughout the layers on two standard classification datasets:
\begin{enumerate}
\item \textbf{Image data}. The \emph{MNIST Dataset} \cite{Letal} consisting of images ($28 \times 28$ pixel greyscale) of 70,000 handwritten digits (60,000 for training and 10,000 for testing). The object is to determine  the  digit (from $\{0,1,2,3,4,5,6,7,8,9\} $) from  the image. So the input vector  has $n_{in}= 28^2 = 784$ entries, and the output vector has  $n_{out}=10$ entries.
\item \textbf{Text data}. The \emph{IMDB  Dataset} \cite{Metal} consisting of 25,000 movie reviews  for training and the same number for testing. The object is to determine the sentiment (positive or negative) from the text. We use preprocessed bag-of-words format provided with the database and remove stopwords (like: an, a, the, this, that, etc) found in the Natural Language Toolkit's corpus, and then use the 4000 most frequently used remaining words in our bag-of-words. So the input vector has $n_{in} = 4000$ and the output vector has $n_{out}=2$.
\end{enumerate}

We consider 6 neural network models: three VPNN variants, one standard model for a control, and two mixed models using features of both:

 \begin{enumerate}
 \item \textbf{VPNN}  The first $L-1$ layers are volume preserving and made up of rotation, permutation, diagonal, and coupled Chebyshev activation sublayers as described in Section \ref{model}, with  the number of rotations in each layer equal to $ 2\lceil \log_2( n_{in} )\rceil $ and the Chebyshev parameter set to  $M=2$.
  \item \textbf{VPNN1.3}  The first $L-1$ layers are volume preserving and made up of rotation, permutation, diagonal, and coupled Chebyshev activation sublayers as described in Section \ref{model}, with  the number of rotations in each layer equal to $ 2\lceil \log_2( n_{in} )\rceil $ and the Chebyshev parameter set to  $M=1.3$.
 \item \textbf{VPNNt} The first $L-1$ layers are volume preserving and made up of rotation, permutation, diagonal, and coupled Chebyshev activation sublayers as described in Section \ref{model}, with   the number of rotations in each layer equal to $2\lceil \log_2( n_{in} )\rceil $ but the Chebyshev parameters are trainable as described in Subsection \ref{Ct}.
 \item \textbf{S-ReLU} The first $L-1$ layers use a standard affine sublayer, $\vx \rightarrow W\vx +\vb$ followed by a ReLU activation function.  (We considered also testing this model with a  sigmoid activation function, however training was problematic due to vanishing gradients.)
 \item \textbf{Mixed1} The first $L-1$ layers use a standard affine sublayer, $\vx \rightarrow W\vx +\vb$ but use coupled Chebyshev activation sublayers with $M=1.3$.
 \item \textbf{Mixed2} The first $L-1$ layers are volume preserving and made up of rotation, permutation and  diagonal  sublayers as described in Section \ref{model}, with the number of rotations in each layer equal to $ 2\lceil \log_2( n_{in} )\rceil $, but the activation function is ReLU.
 \end{enumerate}

 Some specifics of the implementation of the testing are:
 \begin{enumerate}
 \item \textbf{Method:} We use Stochastic Gradient Descent with momentum set to 0.9 and with a batch size of 100 in all training.
 \item \textbf{Layers:} For ease of comparison, all the models we consider  have $L-1$ layers of the same type which preserve dimension (so  the number of neurons in each of  the first $L$ layers is equal to $n_{in}$, the number of input neurons) followed by a fixed downsizer matrix $Z$ as in the basic VPNN model. For testing accuracy we take $L=4$ and for testing learning throughout the layers we take $L=10$.
 \item \textbf{Learning Rate:} The surface of the error function seems to be smoother, and generally less steep for VPNN models than for standard models. This allows us to take larger step sizes (learning rates) at the start of training but causes slower convergence (for the same learning rate) later in training. When testing for accuracy ($L=4$), to accommodate for this and speed up training, we perform a variation of adaptive learning methods: we perform some preliminary runs (with a small number of batches) with larger learning rates to determine stability, and choose initial learning rate of 1/10 of the limit where training seems stable. So for the first half of the training, the learning rates are in the range of 0.1 to 1.0 and then as we have supposedly zeroed in on the minimum, the learning rate is set to 0.01  for all models. When testing for learning throughout the layers ($L=10$), we hold the learning rate at 0.01 for all models.
 \item \textbf{Error Function:}  We use the cross-entropy loss function
\begin{equation} E(\vy, \hat{\vy} ) = -\sum_{i} \hat{\vy}_i \log( \vy_i )
\end{equation}
 (where $\vy$ is the predicted output for input $\vx$ and $\hat{\vy}$ is the actual output) for the error function.
 \end{enumerate}

 \subsection{Testing Accuracy and Training Times on a Four-Layer Neural Network}

 Using a four layer network and running 30 epochs for MNIST and 40 epochs for IMDB, we obtain the training times and accuracy rates as shown in Table  \ref{at}.
 \begin{table}[t]
 \begin{center}
 \begin{tabular}{| c || r | r || r |r ||}
  \hline
  \multirow{2}{*}{Model}
      & \multicolumn{2}{c ||}{MNIST}  &  \multicolumn{2}{c ||}{IMDB } \\
      \cline{2-5}
   &\vbox{\hbox{Training}\hbox{Time}} & Accuracy &\vbox{\hbox{Training}\hbox{Time}}  & Accuracy\\  \hline
  VPNN &29 s/epoch  & 98.06 \%  & 27 s/epoch & 86.89\%  \\
\hline
 VPNN1.3 & 29 s/epoch  & 97.21 \%   & 27 s/epoch&87.46\%   \\
\hline
VPNNt & 29 s/epoch  & 97.38 \%   &27 s/epoch& 83.89\%     \\
\hline
S-ReLU &6 s/epoch  & 97.42 \%   &14 s/epoch& 86.35\%  \\
\hline
Mixed1 & 7 s/epoch  & 98.40 \%  &15 s/epoch& 87.16\%     \\
\hline
Mixed2 & 27 s/epoch  & 96.00 \% &25 s/epoch & 83.90\%   \\
\hline
\end{tabular}
\caption{Training Time and Accuracy} \label{at}
\end{center}
\end{table}

 Fig. \ref{mnist-a} and Fig. \ref{mnist-a-z} (for MNIST) and Fig. \ref{imdb-a} and Fig. \ref{imdb-a-z} (for IMDB) show the progression of the accuracy throughout the training.

Some comments on accuracy:
 \begin{itemize}
 \item  All the models perform comparably well, and very close to the state of the art  for these classication tasks (approximately 99\% for MNIST, and approximately 88\% for IMDB using the best bag-of-words approach). The training times are also comparable, which may be a bit surprising considering all the trigonometric evaluations in the VPNN model.
 \item The  swings in accuracy early in the training are due to the large learning rate. This could obviously be smoothed with a smaller learning rate (and thus more epochs).
 \item VPNN seems to be the  superior volume-preserving neural net, with VPNNt being the least accurate. This is somewhat surprising as there is more ``freedom'' due to additional parameters in the VPNNt model. It may have something to do with the different types of parameters (rotational, bias, diagonal and Chebychev). This seems to cause slow training of the Chebychev parameters in particular. The VPNNt accuracy is still trending upwards after most other models have levelled off. Running significantly more epochs does improve VPNNt performance.
 \item The best model overall for accuracy is Mixed1. This model incorporates some of the best features from both models: the significantly increased parameter space of the standard model, and the gradient control of the coupled Chebychev activation functions of VPNN models.
 \end{itemize}

\begin{figure}[t]
    \centering
    \begin{minipage}{0.47\textwidth}
        \centering
        \includegraphics[width=1.1\textwidth]{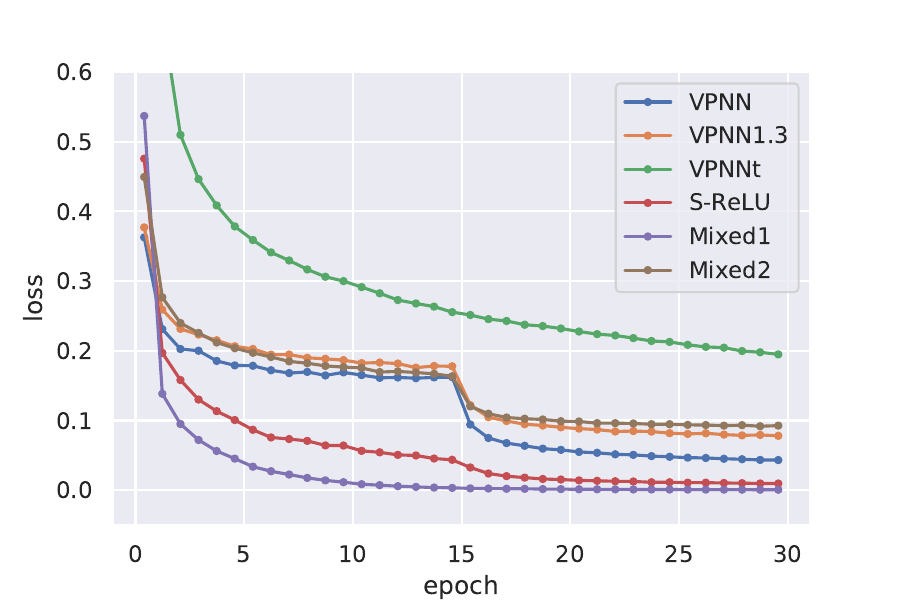} 
        \caption{Error Function: MNIST}\label{mnist-e}
    \end{minipage}\hskip .03\textwidth
    \begin{minipage}{0.47\textwidth}
        \centering
        \includegraphics[width=1.1\textwidth]{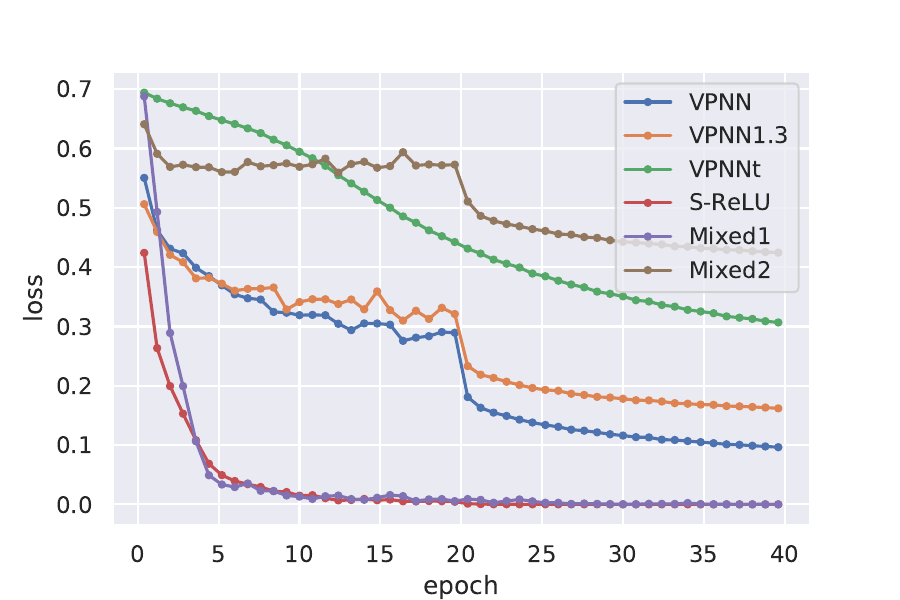} 
        \caption{Error Function: IMDB}\label{imdb-e}
    \end{minipage}
\end{figure}

As Figures \ref{mnist-e} and \ref{imdb-e} show, all models also show similar performance in terms of minimizing the error function as well. Once again, VPNNt is lagging behind but still trending downwards and improves with additional training.

 One factor that should be taken into consideration is the number of parameters in the various models. Fully-connected layer models (like S-ReLU) use   $ w  (w +1)$ parameters per layer of width $w$, versus   $w \left( \lceil \log_2 ( w) \rceil  +2 \right)$
 parameters  per layer VPNN2 and VPNN1.3 (or  $ w \left( \lceil \log_2 ( w) \rceil  +5/2 \right)$   for VPNNt).

 For our models $w= n_{in}$, and Table \ref{para} shows the number of parameters per layer for the different models:

 \begin{table}[t]
 \begin{center}
 \begin{tabular}{| c || r | r |}
  \hline
  Model        & MNIST &  IMDB  \\
       \hline
  VPNN & $9.4\times 10^3$ & $5.6 \times 10^4 $   \\
\hline
 VPNN1.3 & $9.4\times 10^3$ & $5.6 \times 10^4 $  \\
\hline
VPNNt & $9.8\times 10^3$ & $5.8 \times 10^4 $     \\
\hline
S-ReLU & $6.2\times 10^5$ & $1.6 \times 10^7$  \\
\hline
Mixed1 &$6.2\times 10^5$& $1.6 \times 10^7$  \\
\hline
Mixed2 & $9.4\times 10^3$ &  $5.6 \times 10^4 $   \\
\hline
\end{tabular}
\caption{Parameters per layer} \label{para}
\end{center}
\end{table}

 Especially for datasets where each datapoint has a large number of entries, the number of parameters is dramatically lower for VPNNs than for standard neural networks.

 \subsection{Testing  Learning Throughout the Layers on a Ten-Layer  Neural Network}


 We consider the amount of learning throughout the layers for the various models. This will show how well the VPNNs control the gradient in deep neural networks and allow for learning in all layers roughly equally. The magnitude of the vectors $\delta^{(l)}$ are a measure this, as they  indicate how well the parameter  updating has propagated back to the $l$-th layer.  If we have vanishing gradients, we would expect $\| \delta^{(l)} \|$ to be small for early layers ($l$ close to 1) compared to later $\| \delta^{(l)} \|$ ($l$ close to $L$) as the training progresses. If we have exploding gradient we expect the reverse. If all are comparable in size, we have ideal backpropagation.

 For testing learning throughout the layers  we use deeper neural networks. We set $L=10$ layers so there are 9  layers of volume-preserving or standard type, followed by a fixed matrix downsizer output layer.  Since we aren't testing the accuracy here, we run 3 epochs only and collect the norms of the vectors $\delta^{(l)}$ at this stage.

  As it is the comparison of the order of magnitude (rather  than the exact value) of the gradients across the  layers which is relevant, we consider the $\log_{10}$ of the learning amount in each layer compared to  $\log_{10}$ of the learning amount in the final layer for each of the models, so we are plotting
\begin{equation}
 y = \log_{10} \left( \frac{\| \delta^{(l)} \| } {\| \delta^{(L)} \| } \right) = \log_{10} \left(\| \delta^{(l)} \| \right) -  \log_{10} \left( \| \delta^{(L)} \| \right)
\end{equation}
for $l=1,2,3, \ldots, L$. (So, for a given $l$, 10 raised to the corresponding value of $y$ gives the percentage more (or less) of learning in that layer as compared to layer $L$.

\begin{figure}[t]
    \centering
    \begin{minipage}{0.47\textwidth}
        \centering
        \includegraphics[width=1.1\textwidth]{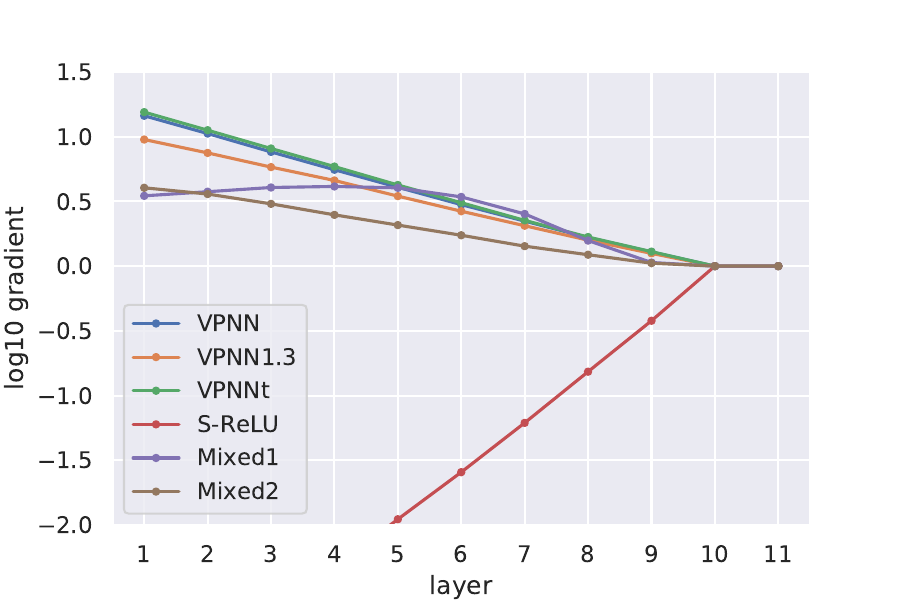} 
        \caption{Learning in the Layers: MNIST}\label{mnist-l}
    \end{minipage}\hskip .03\textwidth
    \begin{minipage}{0.47\textwidth}
        \centering
        \includegraphics[width=1.1\textwidth]{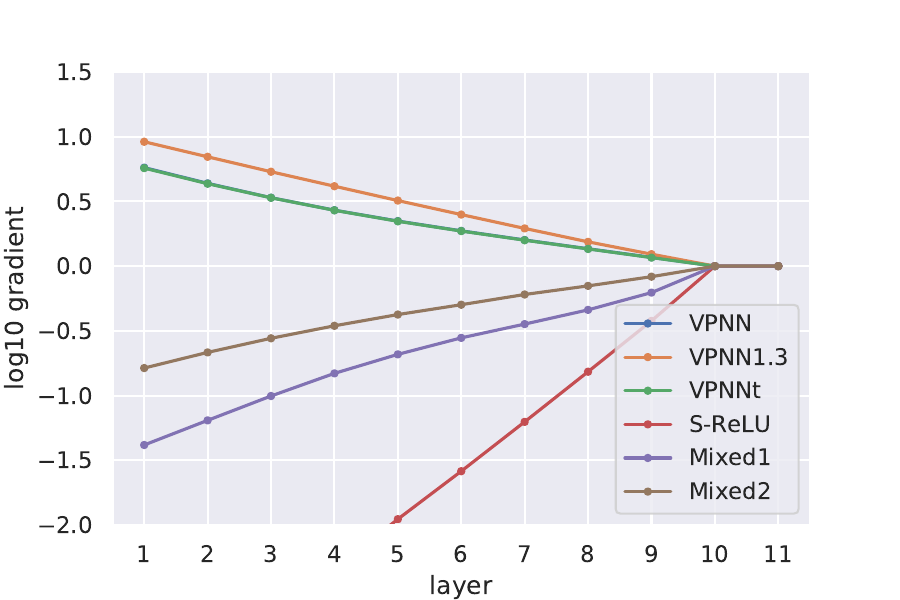} 
        \caption{Learning in the Layers: IMDB}\label{imdb-l}
    \end{minipage}
\end{figure}

  Figures \ref{mnist-l} (for MNIST) and \ref{imdb-l} (for IMDB)  display the data from these runs.

 In Figures \ref{mnist-l} and  \ref{imdb-l}, a positive slope indicates vanishing gradients.
 More precisely, a slope of $m$ on these graphs indicates that learning decreases (when $m$ is positive) or increases (when $m$ is negative) by a factor of $10^{-m}$ for each layer deeper we go into the neural network.
 For S-ReLU in both Figure \ref{mnist-l} and Figure \ref{imdb-l}, the slope is approximately $0.4$ so for every  layer retreat into the network, the gradients (and hence the amount of  learning) decrease by (approximately) a factor of $10^{-0.40} =0.40$. So in layer 1, there is roughly $(0.40)^9 \approx 2.6 \times 10^{-4}$ as much learning as in layer 10. Almost all the learning is in the late layers. Contrast this with VPNN models, where learning is comparable across the layers, and in fact there is slightly more learning in early layers than in late layers. (The learning throughout the layers for VPNN seems to be missing in Figure \ref{imdb-l}, in fact it is basically identical to that for VPNNt, and is obscured by that line.) The mixed models show learning throughout the layers superior to S-ReLU but inferior to the VPNN models.

 The VPNN variants show  clearly superior learning throughout the layers, with no vanishing gradient as compared to standard neural networks.

 While we were not testing accuracy in the ten-layer network, we will mention for interest's sake that the ten-layer VPNNs do train relatively accurately if training is allowed to continue. Since the corresponding four-layer neural networks train close to the standard in both the MNIST and IMDB cases, and we are adding additional complexity, it is perhaps not surprising that accuracy does drop by 5\% to 10\% from the four-layer to ten-layer VPNN in these cases. While VPNN was the most accurate in the four-layer network, VPNN1.3 is the most accurate in the ten-layer case. This is perhaps due to the fact that with additional layers, less ``squashing'' must be done in any particular layer.
 \section{Conclusions} \label{wrap}

We presented a new neural network architecture wherein all layers of the network are volume preserving, this includes a coupled activation layer, however on classification tasks we require a downsizing layer as the last layer. This last layer cannot therefore be volume preserving. We tested the network on two standard classification tasks and while the use of a fully volume-preserving network for classification tasks is unlikely to be the eventual application of this architecture we do showcase the ability of our architecture to both learn and preserve gradients. Demonstrating the preservation of gradients is our main goal in tackling these classification tasks as there are architectures such as convolutional neural networks that perform extremely well on these tasks and especially image classification tasks in general \cite{byerly2020branching, hirata2020ensemble}.

The basic VPNN model here was stripped down to its essentials for the purposes of demonstrating the efficacy of the model. In practice it should be another tool in the machine learning toolbox, used in conjunction with other approaches and techniques to achieve best results. That is one of our main goals for future work. Now that we have the basic model, we plan to consider variations to tackle different applications. Of particular interest for future work will be developing VPNNs into feasible models to handle problems with sequential data with long-term dependencies and generative models. We also note the model may have applications where small memory footprints or smaller numbers of trainable parameters are required or to augment existing architectures.


\subsection{Accessing the code}
The code for the VPNN architecture is accessible in Github repository (\cite{GitVPNN}).

\bibliographystyle{IEEEtran}
\bibliography{VPNNarXiv}

\begin{thebibliography}{10}
\providecommand{\url}[1]{#1}
\csname url@samestyle\endcsname
\providecommand{\newblock}{\relax}
\providecommand{\bibinfo}[2]{#2}
\providecommand{\BIBentrySTDinterwordspacing}{\spaceskip=0pt\relax}
\providecommand{\BIBentryALTinterwordstretchfactor}{4}
\providecommand{\BIBentryALTinterwordspacing}{\spaceskip=\fontdimen2\font plus
\BIBentryALTinterwordstretchfactor\fontdimen3\font minus
  \fontdimen4\font\relax}
\providecommand{\BIBforeignlanguage}[2]{{%
\expandafter\ifx\csname l@#1\endcsname\relax
\typeout{** WARNING: IEEEtran.bst: No hyphenation pattern has been}%
\typeout{** loaded for the language `#1'. Using the pattern for}%
\typeout{** the default language instead.}%
\else
\language=\csname l@#1\endcsname
\fi
#2}}
\providecommand{\BIBdecl}{\relax}
\BIBdecl

\bibitem{S2}
J.~Schmidhuber, ``Deep learning in neural networks: An overview,'' \emph{Neural
  networks}, vol.~61, pp. 85--117, 2015.

\bibitem{KSH}
A.~Krizhevsky, I.~Sutskever, and G.~E. Hinton, ``Imagenet classification with
  deep convolutional neural networks,'' in \emph{Advances in neural information
  processing systems}, 2012, pp. 1097--1105.

\bibitem{Hetal}
G.~Hinton, L.~Deng, D.~Yu, G.~Dahl, A.-r. Mohamed, N.~Jaitly, A.~Senior,
  V.~Vanhoucke, P.~Nguyen, B.~Kingsbury \emph{et~al.}, ``Deep neural networks
  for acoustic modeling in speech recognition,'' \emph{IEEE Signal processing
  magazine}, vol.~29, 2012.

\bibitem{Cetal}
K.~Cho, B.~Van~Merri{\"e}nboer, C.~Gulcehre, D.~Bahdanau, F.~Bougares,
  H.~Schwenk, and Y.~Bengio, ``Learning phrase representations using {RNN}
  encoder-decoder for statistical machine translation,'' \emph{arXiv preprint
  arXiv:1406.1078}, 2014.

\bibitem{H}
S.~Hochreiter, ``Untersuchungen zu dynamischen neuronalen netzen,''
  \emph{Diploma, Technische Universit{\"a}t M{\"u}nchen}, vol.~91, no.~1, 1991.

\bibitem{BSF}
Y.~Bengio, P.~Simard, P.~Frasconi \emph{et~al.}, ``Learning long-term
  dependencies with gradient descent is difficult,'' \emph{IEEE Transactions on
  Neural Networks}, vol.~5, no.~2, pp. 157--166, 1994.

\bibitem{glorot2010understanding}
X.~Glorot and Y.~Bengio, ``Understanding the difficulty of training deep
  feedforward neural networks,'' in \emph{Proceedings of the thirteenth
  international conference on artificial intelligence and statistics}.\hskip
  1em plus 0.5em minus 0.4em\relax JMLR Workshop and Conference Proceedings,
  2010, pp. 249--256.

\bibitem{nair2010rectified}
V.~Nair and G.~E. Hinton, ``Rectified linear units improve restricted
  {B}oltzmann machines,'' in \emph{Proceedings of the 27th international
  conference on machine learning (ICML-10)}, 2010, pp. 807--814.

\bibitem{pascanu2013difficulty}
R.~Pascanu, T.~Mikolov, and Y.~Bengio, ``On the difficulty of training
  recurrent neural networks,'' in \emph{International conference on machine
  learning}.\hskip 1em plus 0.5em minus 0.4em\relax PMLR, 2013, pp. 1310--1318.

\bibitem{HS}
S.~Hochreiter and J.~Schmidhuber, ``Long short-term memory,'' \emph{Neural
  computation}, vol.~9, no.~8, pp. 1735--1780, 1997.

\bibitem{S1}
J.~Schmidhuber, ``Learning complex, extended sequences using the principle of
  history compression,'' \emph{Neural Computation}, vol.~4, no.~2, pp.
  234--242, 1992.

\bibitem{saxe2013exact}
A.~M. Saxe, J.~L. McClelland, and S.~Ganguli, ``Exact solutions to the
  nonlinear dynamics of learning in deep linear neural networks,'' \emph{arXiv
  preprint arXiv:1312.6120}, 2013.

\bibitem{bansal2018can}
N.~Bansal, X.~Chen, and Z.~Wang, ``Can we gain more from orthogonality
  regularizations in training deep cnns?'' \emph{arXiv preprint
  arXiv:1810.09102}, 2018.

\bibitem{henaff2016recurrent}
M.~Henaff, A.~Szlam, and Y.~LeCun, ``Recurrent orthogonal networks and
  long-memory tasks,'' in \emph{International Conference on Machine
  Learning}.\hskip 1em plus 0.5em minus 0.4em\relax PMLR, 2016, pp. 2034--2042.

\bibitem{arjovsky2016unitary}
M.~Arjovsky, A.~Shah, and Y.~Bengio, ``Unitary evolution recurrent neural
  networks,'' in \emph{International Conference on Machine Learning}.\hskip 1em
  plus 0.5em minus 0.4em\relax PMLR, 2016, pp. 1120--1128.

\bibitem{wisdom2016full}
S.~Wisdom, T.~Powers, J.~R. Hershey, J.~L. Roux, and L.~Atlas, ``Full-capacity
  unitary recurrent neural networks,'' \emph{arXiv preprint arXiv:1611.00035},
  2016.

\bibitem{GBB}
X.~Glorot, A.~Bordes, and Y.~Bengio, ``Deep sparse rectifier neural networks,''
  in \emph{Proceedings of the fourteenth international conference on artificial
  intelligence and statistics}, 2011, pp. 315--323.

\bibitem{Letal}
Y.~LeCun \emph{et~al.}, ``The {MNIST} dataset of handwritten digits (images),''
  1999.

\bibitem{Metal}
A.~L. Maas, R.~E. Daly, P.~T. Pham, D.~Huang, A.~Y. Ng, and C.~Potts,
  ``Learning word vectors for sentiment analysis,'' in \emph{Proceedings of the
  49th annual meeting of the association for computational linguistics: Human
  language technologies-volume 1}.\hskip 1em plus 0.5em minus 0.4em\relax
  Association for Computational Linguistics, 2011, pp. 142--150.

\bibitem{byerly2020branching}
A.~Byerly, T.~Kalganova, and I.~Dear, ``A branching and merging convolutional
  network with homogeneous filter capsules,'' \emph{arXiv preprint
  arXiv:2001.09136}, 2020.

\bibitem{hirata2020ensemble}
D.~Hirata and N.~Takahashi, ``Ensemble learning in cnn augmented with fully
  connected subnetworks,'' \emph{arXiv preprint arXiv:2003.08562}, 2020.

\bibitem{GitVPNN}
G.~MacDonald, A.~Godbout, B.~Gillcash, and S.~Cairns, ``Python ({P}ytorch) code
  for {VPNN},'' \emph{\url{https://github.com/andrewgodbout/VPNN\_pytorch}},
  2019.

\end{thebibliography}

\end{document}